\documentclass{ecai} 

\usepackage{latexsym}
\usepackage{multirow}
\usepackage{bm}
\usepackage{amssymb}
\usepackage{amsmath}
\usepackage{amsthm}
\usepackage{booktabs}
\usepackage{enumitem}
\usepackage{graphicx}
\usepackage{color}

\usepackage{subcaption}
\usepackage{cite, url}
\usepackage{amsfonts}
\usepackage{algorithmic}
\usepackage{textcomp}
\usepackage{xcolor}
\usepackage{threeparttable}
\usepackage{caption} 
\usepackage{float}

\definecolor{GrassGreen}{RGB}{146,208,80}
\definecolor{SkyBlue}{RGB}{4,175,252}
\definecolor{FluoGreen}{RGB}{24,229,25}

\begin{document}


\begin{frontmatter}




\title{MoSt-DSA: Modeling Motion and Structural Interactions for Direct Multi-Frame Interpolation in DSA Images}


\author[A]{\fnms{Ziyang}~\snm{Xu}}
\author[B]{\fnms{Huangxuan}~\snm{Zhao}}
\author[A]{\fnms{Ziwei}~\snm{Cui}}
\author[A]{\fnms{Wenyu}~\snm{Liu}}
\author[B]{\fnms{Chuansheng}~\snm{Zheng}}
\author[A]{\fnms{Xinggang}~\snm{Wang}\thanks{Corresponding Author. Email: xgwang@hust.edu.cn}} 

\address[A]{Institute of AI, School of EIC, Huazhong University of Science and Technology}
\address[B]{Union Hospital, Tongji Medical College, Huazhong University of Science and Technology}


\begin{abstract}
Artificial intelligence has become a crucial tool for medical image analysis. As an advanced cerebral angiography technique, Digital Subtraction Angiography (DSA) poses a challenge where the radiation dose to humans is proportional to the image count. By reducing images and using AI interpolation instead, the radiation can be cut significantly. However, DSA images present more complex motion and structural features than natural scenes, making interpolation more challenging. We propose MoSt-DSA, the first work that uses deep learning for DSA frame interpolation. Unlike natural scene Video Frame Interpolation (VFI) methods that extract unclear or coarse-grained features, we devise a general module that models motion and structural context interactions between frames in an efficient full convolution manner by adjusting optimal context range and transforming contexts into linear functions. Benefiting from this, MoSt-DSA is also the first method that directly achieves any number of interpolations at any time steps with just one forward pass during both training and testing. We conduct extensive comparisons with 7 representative VFI models for interpolating 1 to 3 frames, MoSt-DSA demonstrates robust results across 470 DSA image sequences (each typically 152 images), with average SSIM over 0.93, average PSNR over 38 (standard deviations of less than 0.030 and 3.6, respectively), comprehensively achieving state-of-the-art performance in accuracy, speed, visual effect, and memory usage. Our code is available at https://github.com/ZyoungXu/MoSt-DSA.
\end{abstract}

\end{frontmatter}

\section{Introduction}

Frame interpolation, a class of fundamental tasks in computer vision, aims to deduce intermediate frames from given preceding and succeeding ones \cite{huang2020rife,lu2022video}. These tasks are classified into single-frame and multi-frame interpolation based on the number of frames inferred \cite{Shang_2023_CVPR,jiang2018super,Kalluri_2023_WACV}. Traditionally, multi-frame interpolation is achieved recursively. For instance, an intermediate frame $\bm{I}_{b}$ is inferred first, and then used with the ground truths of adjacent frames to deduce additional frames $\bm{I}_{a}$ and $\bm{I}_{c}$ \cite{reda2022film,niklaus2020softmax,park2021abme}. However, this approach neither supports direct multi-frame interpolation nor allows flexible determination of frame count (typically odd).

DSA is an advanced medical imaging technology widely used in interventional surgery \cite{radiation2040028}. It is crucial for diagnosing and treating various vascular diseases, including brain, heart, and limbs. DSA operates by injecting a contrast agent, usually iodine-based, into the patient and capturing vascular images with X-rays. DSA technology varies: 2D DSA provides basic two-dimensional images. 3D DSA, capturing images from multiple angles \cite{zhao2022self}. 4D DSA adds a time dimension, forming a sequence that captures dynamic blood flow's changes over time \cite{Ito20224d}.

\begin{figure}[t]
    \centering
    \includegraphics[width=0.9\linewidth]{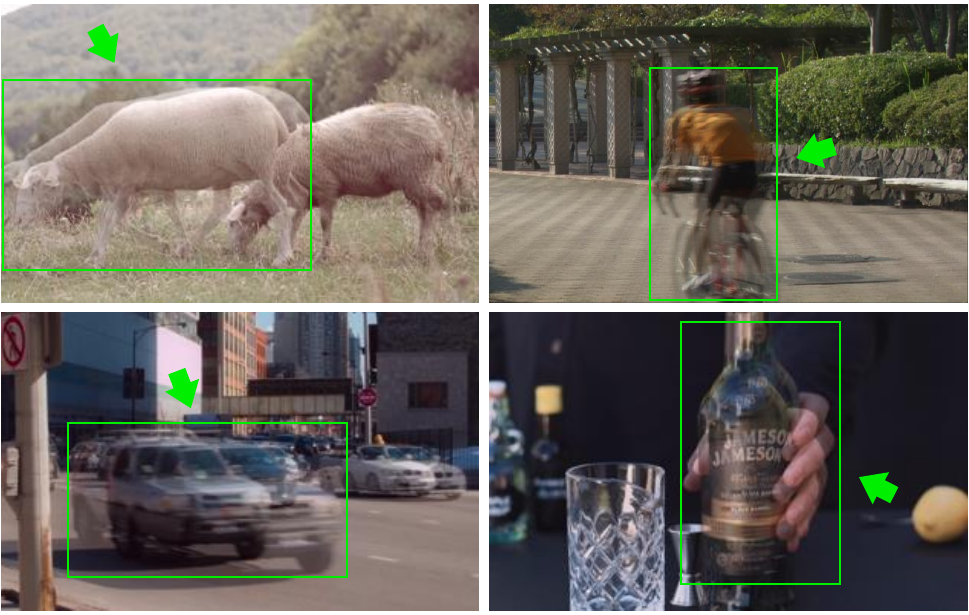}
    \caption{\textbf{Various motions in natural scenes.} Motion subjects have simple structure and coarse texture feature granularity, also easy to predict the motion trajectory.}
    \label{fig:natural_image}
\end{figure}

\begin{figure}[t]
    \centering
    \includegraphics[width=0.9\linewidth]{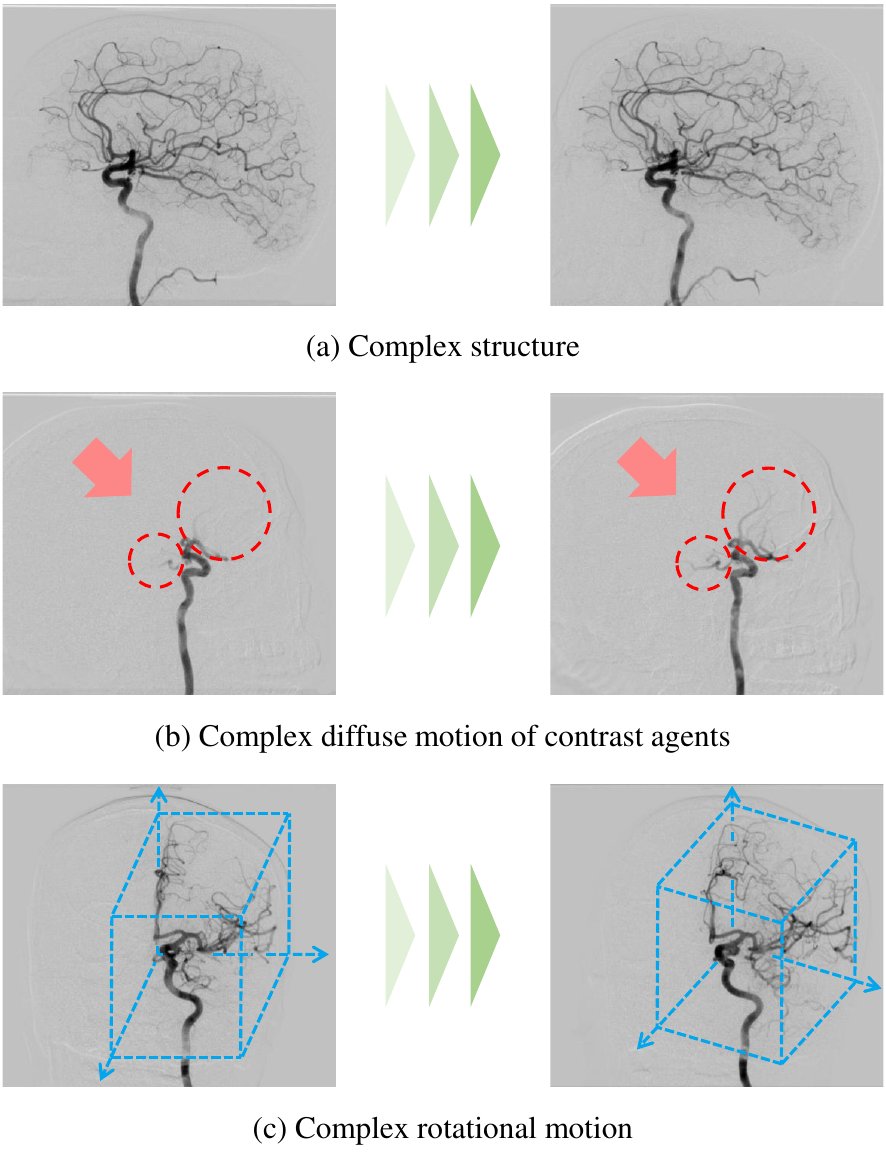}
    \caption{Various challenges for frame interpolation in DSA images.}
    \label{fig:intro}
\end{figure}

Moreover, frame interpolation for DSA images differs significantly from natural images. As comparing Fig. \ref{fig:natural_image} with Fig. \ref{fig:intro}, DSA images present more complex structural and motion details \cite{Guo_2020_CVPR}. Currently, no specific interpolation solutions for DSA images exist.

\begin{figure}[t]
    \centering
    \includegraphics[width=\linewidth]{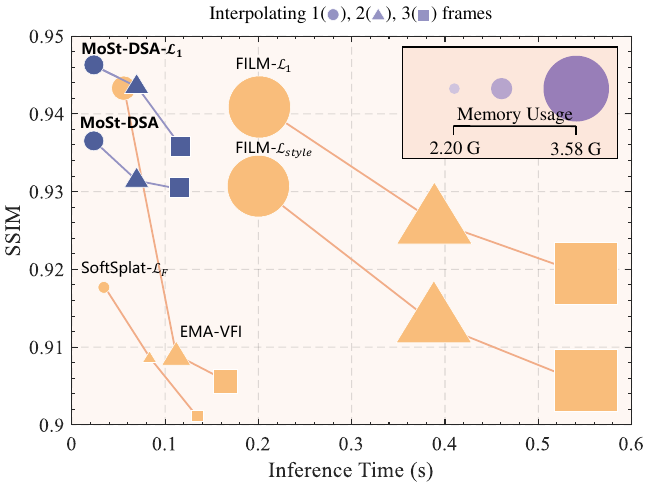}
    \caption{\textbf{SSIM-Time-Memory comparison of different methods for direct interpolating 1 to 3 frames on our DSA dataset.} Our MoSt-DSA-$\mathcal{L}_1$ achieves 94.62, 94.35, 93.58 SSIM, 0.024s, 0.070s, 0.117s inference time, and 2.59G, 2.61G, 2.61G memory usage for interpolating 1 to 3 frames, respectively, outperforming SOTA EMA-VFI \protect\cite{zhang2023extracting} in all aspects. Details in Tab. \ref{tab:inf1_ave},\ref{tab:inf2_ave},\ref{tab:inf3_ave}.}
    \label{fig:coverchart}
\end{figure}

As we move from 2D to 4D DSA, frame interpolation complexity increases.
Our research targets direct multi-frame interpolation for 4D DSA.
Hereafter, DSA refers specifically to 4D DSA unless stated otherwise. Frame interpolation for DSA images confronts challenges from complex structures and motions. First, the vascular structure is complex: vessels are irregular, dense, and varied in size, like Fig. \ref{fig:intro}(a). Second, the imaging captures the contrast agent's diffusion, a non-rigid and complex motion depicted in Fig. \ref{fig:intro}(b). Third, vessels rotate during imaging, causing occlusions and overlaps that complicate motion analysis, as shown in Fig. \ref{fig:intro}(c).

\begin{figure}[t]
    \centering
    \includegraphics[width=0.95\linewidth]{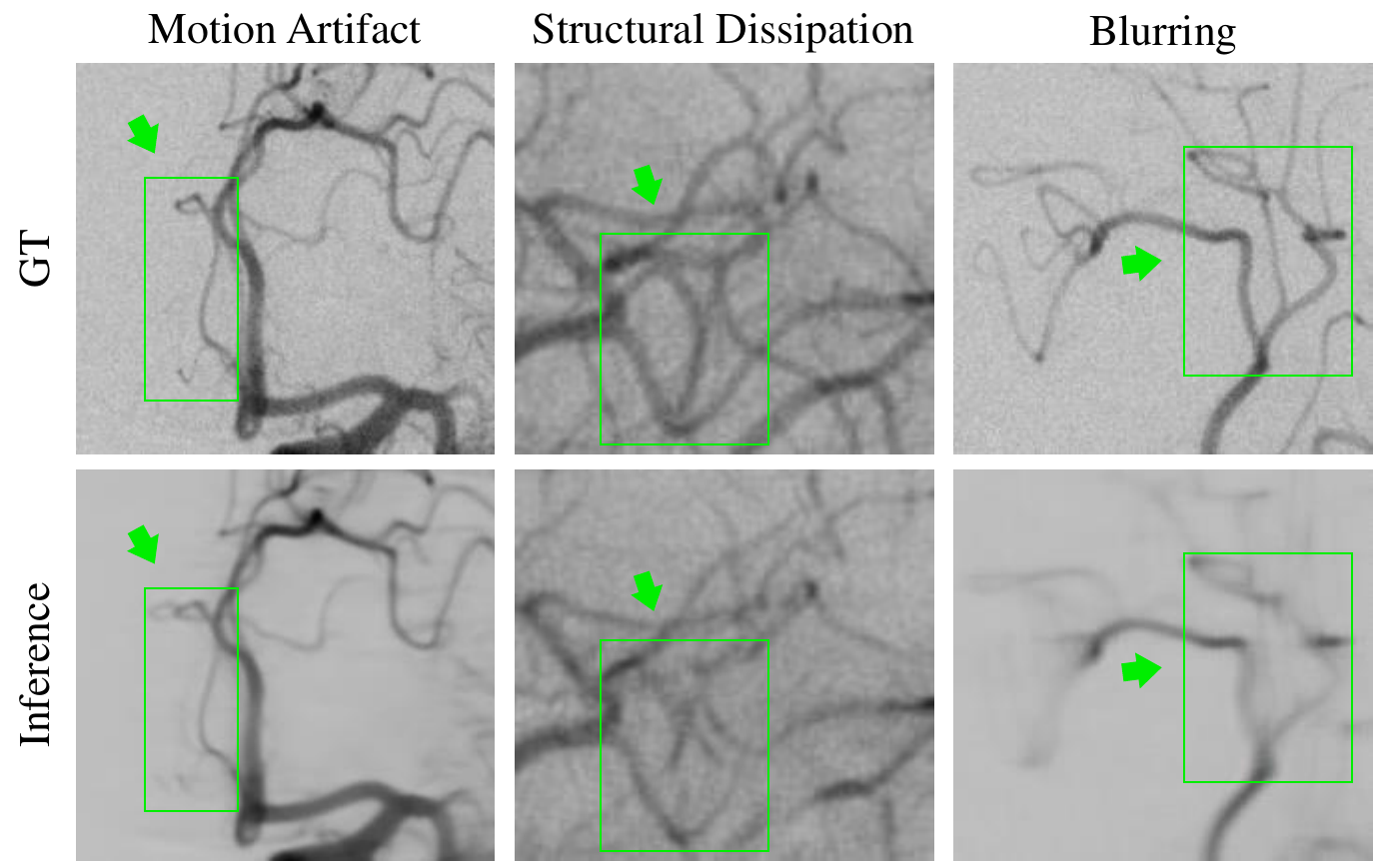}
    \caption{Existing frame interpolation methods are tailored for natural scenes, and commonly exhibit issues such as motion artifacts, structural dissipation, and blurring in DSA frame interpolation.}
    \label{fig:problems}
\end{figure}

To address the above challenges, extracting fine-grained and precise motion and structural features is critical. However, existing frame interpolation methods are tailored for natural scenes, resulting in unclear or coarse extraction of motion and structural features for DSA images. Common approaches fall into three categories. The first uses a single module to mix and extract both motion and structural features, resulting in ambiguity in both aspects\cite{Kalluri_2023_WACV,kong2022ifrnet,lu2022video,bao2019depth,ding2021cdfi}. The second designs multiple modules to sequentially extract structural features of each frame and motion features between frames, although clear motion features are obtained, the corresponding structural relationships between frames are lacking\cite{gao2023video,zhou2023video,yu2023range,danier2022st,jia2022neighbor,niklaus2018context,park2021abme,reda2022film,sim2021xvfi,xue2019video}. The third designs a single module to extract relative motion and structural features from frames simultaneously, but due to coarse context granularity, it fails to adapt to the fine-grained, complex structures of DSA images\cite{zhang2023extracting}. These methods commonly exhibit issues such as motion artifacts, structural dissipation, and blurring in DSA frame interpolation, as shown in Fig. \ref{fig:problems}.

In this work, we propose a network for flexible, efficient, and direct multi-frame interpolation in DSA images. Initially, we extract multi-scale features from input frames and enhance them through cross-scale fusion. Inspired by the EMA-VFI \cite{zhang2023extracting}, we propose a general module named MSFE that extracts motion and structural features between enhanced frames by cross-attention. Unlike EMA-VFI, MSFE doesn't rely on expensive attention maps and can flexibly adjust context-aware granularity. Specifically, by adjusting the optimal context range and transforming contexts into linear functions, MSFE calculates cross-attention between input frames in a fully convolutional manner, which reduces the storage cost and increases the computing speed. After extracting general motion and structural features through MSFE, we map the motion features at different times $t$ and decode them together with the structural features to obtain flows and masks. Finally, a simplified UNet \cite{ronneberger2015u} refines features at different scales, decoding the flows, masks, and structural features to produce the corresponding intermediate frame $\bm{I}_{t}$. A key advantage is that by extracting general motion and structural features only once, our MoSt-DSA can interpolate any intermediate frame by combining different time $t$ during both training and testing. This is more flexible than methods interpolating for fixed $t$ \cite{park2021abme}, more efficient than methods extracting different features for multiple $t$ \cite{Kalluri_2023_WACV}, and more direct than methods interpolating multi frames recursively \cite{reda2022film,niklaus2020softmax}.

In summary, our work offers these main contributions:
\begin{itemize}
\item To our knowledge, MoSt-DSA is the first work that uses deep learning for DSA frame interpolation, and also the first method that directly achieves any number of interpolations at any time steps with just one forward pass during both training and testing.
\item We propose a general module named MSFE that models motion and structural context interactions between frames by cross-attention. Significantly, by adjusting the optimal context range and transforming contexts into linear functions, MSFE calculates cross-attention in a fully convolutional manner, which reduces the storage cost and increases the computing speed.
\item We conduct extensive comparisons with 7 representative VFI models for interpolating 1 to 3 frames, MoSt-DSA demonstrates robust results across 470 DSA image sequences (each typically 152 images), with average SSIM over 0.93, average PSNR over 38 (standard deviations of less than 0.030 and 3.6, respectively), comprehensively achieving state-of-the-art performance in accuracy, speed, visual effect, and memory usage. If applied clinically, MoSt-DSA can significantly reduce the DSA radiation dose received by doctors and patients, lowering it by 50\%, 67\%, and 75\% when interpolating 1 to 3 frames, respectively.
\end{itemize}

\section{Related Work}

\subsection{Direct Multi-Frame Interpolation}
Interpolation tasks for continuous image sequences, known as Video Frame Interpolation (VFI), aim to generate one or multiple intermediate frames between input frames \cite{Kalluri_2023_WACV,jiang2018super}. Considering that in DSA imaging, where the radiation dose correlates with image count, reducing frames and using AI interpolation instead can cut radiation significantly. Further, if multi-frame interpolation could be rapidly achieved with just one forward pass, it would not only further reduce radiation dose but also shorten the time consumed, securing more precious time for patient treatment. However, advanced VFI methods primarily focused on single-frame interpolation, with multi-frame interpolation often reliant on recursion \cite{zhang2023extracting,reda2022film,niklaus2020softmax,park2021abme}. During training, these methods are limited as they neither directly complete multi-frame interpolation nor allow flexible frame number determination, leading to a significant decrease in accuracy for direct multi-frame interpolation during testing. Unlike these methods, our MoSt-DSA can directly achieve any number of frame interpolations at any time steps with just one forward pass during both training and testing.

\subsection{Modeling Motion and Structural Interactions}

Modeling the motion and structural interactions is essential for extracting motion and structural features. Existing frame interpolation methods are tailored for natural scenes, and the modeling of motion and structural interactions could divided into three categories. The first category concatenates input frames to a backbone network that extracts mixed features of motion and structure \cite{Kalluri_2023_WACV,kong2022ifrnet,lu2022video,bao2019depth,ding2021cdfi}. While straightforward to implement, these methods lack clear motion information, leading to restrictions in interpolating frames with various numbers and time steps \cite{shi2022video,gui2020featureflow,huang2020rife}. The second category utilizes multiple modules to sequentially extract the structural features of each frame and the motion features between frames \cite{gao2023video,zhou2023video,yu2023range,danier2022st,jia2022neighbor,niklaus2018context,park2021abme,reda2022film,sim2021xvfi,xue2019video}. Although these methods provide explicit motion features, they require modules with high computational costs, such as cost volume \cite{jia2022neighbor,park2021abme,reda2022film}. Moreover, capturing structural features from individual frames does not adequately identify the structural correspondence between frames, a critical aspect noted by \cite{jia2022neighbor} for VFI tasks. The third category, represented by \cite{zhang2023extracting}, utilizes a single module for concurrent extraction of relative motion and structural features from frames. This approach's advantages include preserving and enhancing the detailed structural features of input frames without interference from motion features, mapping motion features to any moment for arbitrary intermediate frame generation, and significantly lowering training costs. However, due to coarse context granularity, it fails to adapt to the fine-grained, complex structures of DSA images. These methods commonly exhibit issues such as motion artifacts, structural dissipation, and blurring in DSA frame interpolation, as shown in Fig. \ref{fig:problems}. Our method, aligning with the third category, introduces a general module named MSFE that models motion and structural context interactions between frames by cross-attention. Differing from \cite{zhang2023extracting}, MSFE doesn't rely on expensive attention maps and can flexibly adjust context-aware granularity. By adjusting the optimal context range and transforming available contexts into linear functions, MSFE calculates cross-attention in a fully convolutional manner, which further reduces the storage cost and increases the computing speed.

\begin{figure}[htbp]
    \centering
    \includegraphics[width = 1.0\linewidth]{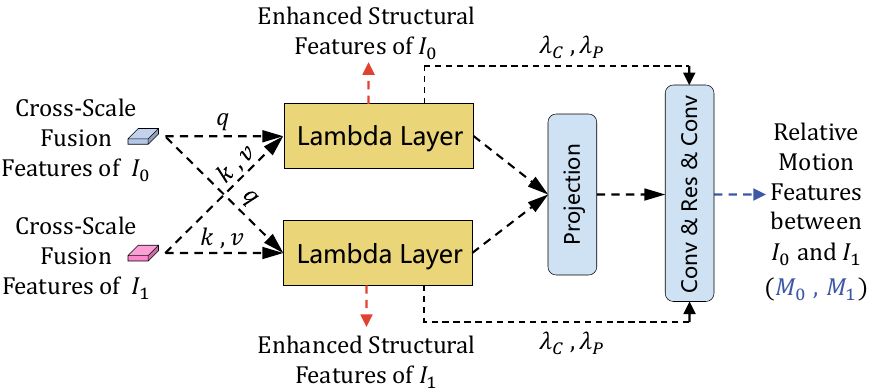}
    \caption{\textbf{An illustration of the MoSt Attention in the MSFE module for calculating motion and structural features.} The enhanced structural features of ${I_0}$ and ${I_1}$ are involved in subsequent calculations in the MSFE module to generate the final structural features, see Fig. \ref{fig:arch} for details.}
    \label{fig:MoSt-Attention}
\end{figure}

\begin{figure}[t]
    \centering
    \includegraphics[width = 0.7\linewidth]{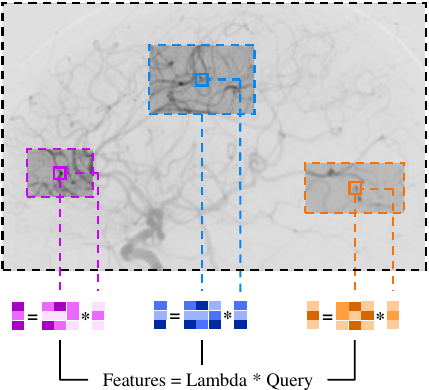}
    \caption{\textbf{An illustration of how we use Lambda Layer to calculate features in our MoSt Attention.} Lambda Layer summarizes contextual information (within a scope $r$) into a fixed-size linear function (i.e. a matrix) applied to the corresponding query, thus bypassing the need for memory-intensive attention maps.}
    \label{fig:lambda}
\end{figure}

\begin{figure*}[t]
    \centering
    \includegraphics[width = 0.9\linewidth]{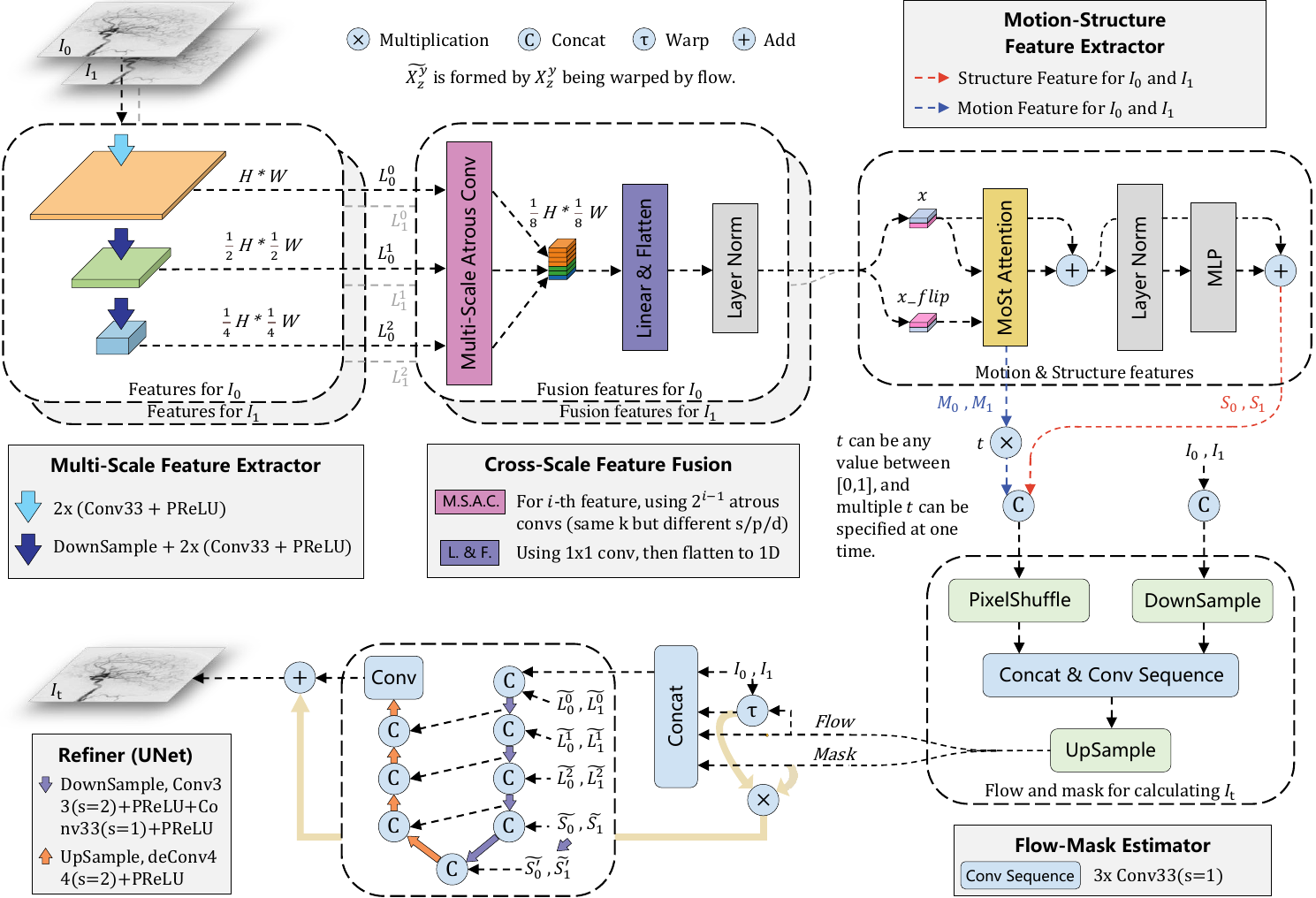}
    \caption{\textbf{Overall network architecture of our MoSt-DSA.} First, the Multi-Scale Feature Extractor (\textbf{FE}) processes the input frames $\bm{I}_{0}$ and $\bm{I}_{1}$ to obtain features at three different scales through continuous convolution and down-sampling. Next, Cross-Scale Feature Fusion (\textbf{CSFF}) uses multi-scale atrous convolutions \protect\cite{chen2017deeplab} to generate fused features for $\bm{I}_{0}$ and $\bm{I}_{1}$, which are linearly mapped and normalized. The Motion-Structure Feature Extractor (\textbf{MSFE}) then calculates motion and structural features from these fused features. Subsequently, for the intermediate time $t$, motion features are mapped. These mapped motion features, along with structural features and the original frames, feed into the Flow-Mask Estimator (\textbf{FME}) to predict flow and mask. Finally, the \textbf{Refiner} combines the various scale features from FE and structural features from MSFE, along with flow and mask, refining them into the image of intermediate time $t$.}
    \label{fig:arch}
\end{figure*}

\section{Method}
Presenting a groundbreaking approach to direct multi-frame interpolation in DSA images, Fig. \ref{fig:arch} delineates the overall network architecture of our method.
Briefly, it is divided into five key modules. Initially, it employs the Multi-Scale Feature Extractor (\textbf{FE}), Cross-Scale Feature Fusion (\textbf{CSFF}), and Motion-Structure Feature Extractor (\textbf{MSFE}) to extract general motion and structural features. Subsequently, the Flow-Mask Estimator (\textbf{FME}) and \textbf{Refiner} decode and refine these features for different moments $t$ to generate frame $I_t$. 

\subsection{Extracting General Motion and Structural Features}

\textbf{Multi-Scale Feature Extractor (FE).} To excavate foundational features of blood vessels of various sizes before extracting motion and structural features, we first employ the FE to derive three different scales of neurovascular features. For input frames $\bm{I}_{0}$ and $\bm{I}_{1}$, we initially compute the third layer of low-level features $\bm{L}_{0}^{0}$ and $\bm{L}_{1}^{0}$, respectively, using 3x3 convolutions followed by PReLU \cite{He_2015_ICCV}. Subsequently, through downsampling and the same convolution and activation configuration, we calculate the second layer of low-level features $\bm{L}_{0}^{1}$ and $\bm{L}_{1}^{1}$, as well as the first layer of low-level features $\bm{L}_{0}^{2}$ and $\bm{L}_{1}^{2}$ for $\bm{I}_{0}$ and $\bm{I}_{1}$ respectively. Mathematically,

\begin{align}
\left\{\begin{array}{l}
\bm{L}_j^0=\bm{H}\left(\bm{I}_j\right) \\
\bm{L}_j^1=\bm{D}\left(\bm{L}_j^0\right) \\
\bm{L}_j^2=\bm{D}\left(\bm{L}_j^1\right)
\end{array}\right.,
\end{align}
where $\bm{H}$ is a stack of convolution and activation functions, while $\bm{D}$ represents an integration of $\bm{H}$ with an additional downsampling operation, and $j$ is 0 or 1.

\textbf{Cross-Scale Feature Fusion (CSFF).} To fuse neurovascular features of different scales and enhance the representation of foundational features, we further employ the CSFF for $\bm{I}_{0}$ and $\bm{I}_{1}$ to perform cross-scale feature fusion. Specifically, for the $i$-th layer low-level features $\bm{L}_{0}^{i}$ and $\bm{L}_{1}^{i}$, we use $2^{i-1}$ atrous convolutions \cite{chen2017deeplab} (with a fixed kernel size of 3, stride of $2^{i}$, and for the $n$-th atrous convolution, both padding and dilation size are $n$). Mathematically,

\begin{align}
\bm{F}\left(\bm{L}_j^i\right)=\left(\bm{A}_1\left(\bm{L}_j^i\right), \ldots, \bm{A}_n\left(\bm{L}_j^i\right)\right),
\end{align}
where $\bm{F}$ signifies feature fusion, $\bm{A}$ indicates atrous convolution. The variable $n$, representing the number of atrous convolutions, takes a value of $2^{i-1}$ for $i$ equal to 0, 1, or 2. Moreover, by merging the fused features from various scales and implementing a linear mapping, we obtain the cross-scale fused features $\bm{F}_0$ and $\bm{F}_1$ for $\bm{I}_0$ and $\bm{I}_1$ respectively, as:

\begin{align}
\bm{F}_j=\mathcal{T}\left[\bm{\mathcal{C}}\left(\bm{F}\left(\bm{L}_j^0\right), \bm{F}\left(\bm{L}_j^1\right), \bm{F}\left(\bm{L}_j^2\right)\right)\right],
\end{align}
where $\bm{\mathcal{T}}$ represents the linear mapping, with $\bm{\mathcal{C}}$ indicates the concatenation operation. Finally, we flatten $\bm{F}_j$ and then normalize it, preparing for subsequent processing by the MSFE.

\textbf{Motion-Structure Feature Extractor (MSFE).} We propose MoSt Attention to calculate relative motion features while enhancing structural features between frames, as shown in Fig. \ref{fig:MoSt-Attention} and \ref{fig:lambda} in detail. To facilitate and simplify understanding, the following formulas we give is based on our actual code implementation. We first concatenate $\bm{F}_0$ and $\bm{F}_1$ to obtain $\bm{F}_a \in \mathbb{R}^{|n| \times d}$, and then acquire $\bm{F}_{a}^{\prime} \in \mathbb{R}^{|n| \times d}$ through reverse concatenation, as:

\begin{align}
\left\{\begin{array}{l}
\bm{F}_a=\bm{\mathcal{C}}\left(\bm{F}_0, \bm{F}_1\right) \\
\bm{F}_{a}^{\prime}=\bm{\mathcal{C}}\left(\bm{F}_1, \bm{F}_0\right)
\end{array}\right..
\end{align}

Furthermore, we employ a Lambda Layer \cite{bello2021lambdanetworks} to simulate content-based and position-based contextual interactions in a fully convolutional manner. Specifically, we denote the depth of query and value as $|k|$ and $|v|$, respectively, and denote the position information with $\bm{P} \in \mathbb{R}^{|n| \times d}$. The queries, keys, and values are calculated as follows:

\begin{align}
\left\{\begin{array}{l}
\bm{Q}=\bm{F}_a \bm{W}_Q \in \mathbb{R}^{|n| \times|k|} \\
\bm{K}=\bm{F}_{a}^{\prime} \bm{W}_K \in \mathbb{R}^{|n| \times|k|} \\
\bm{V}=\bm{F}_{a}^{\prime} \bm{W}_V \in \mathbb{R}^{|n| \times|v|}
\end{array}\right..
\end{align}

Then we represent relative position embeddings as $\bm{E} \in \mathbb{R}^{|n| \times|k|}$. By normalizing the keys, we obtain $\bar{\bm{K}}=\operatorname{softmax}(\bm{K}$, axis $=n)$. Next, we compute the content-based contextual interactions $\bm{\lambda}_c$ and position-based contextual interactions $\bm{\lambda}_p$, as:

\begin{align}
\left\{\begin{array}{l}
\bm{\lambda}_c=\bar{\bm{K}}^T \bm{V} \in \mathbb{R}^{|k| \times|v|} \\
\bm{\lambda}_p=\bm{E}^T \bm{V} \in \mathbb{R}^{|k| \times|v|}
\end{array}\right..
\end{align}

Finally, by applying contextual interactions to the queries as well as $\bm{P}$, we obtain the general motion and structural features necessary for inferring any intermediate frame, as:

\begin{align}
\left\{\begin{array}{l}
\bm{S}=\bm{Q} \bm{\lambda}_c+\bm{Q} \bm{\lambda}_p=\bm{\mathcal{C}}\left(\bm{S}_0, \bm{S}_1\right) \\
\bm{M}=\bm{P} \bm{\lambda}_c+\bm{P} \bm{\lambda}_p=\bm{\mathcal{C}}\left(\bm{M}_0, \bm{M}_1\right)
\end{array}\right..
\end{align}

\subsection{Decoding and Refining for Multi Intermediate Frames}
To further obtain motion features corresponding to multiple different intermediate times $t$, we multiply each $t$ with the general motion features to map and obtain $\bm{M}_{0 \rightarrow t}$ and $\bm{M}_{1 \rightarrow t}$.

\textbf{Flow-Mask Estimator (FME).} For a specific $t$, $\bm{M}_{0 \rightarrow t}$ and $\bm{M}_{1 \rightarrow t}$ are concatenated with $\bm{S}_0$ and $\bm{S}_1$, respectively, and this combination serves as part \emph{a} of the input for the FME, while part \emph{b} is the concatenation of $\bm{I}_0$ and $\bm{I}_1$. As shown in Fig. \ref{fig:arch}, FME (denote by $\bm{\mathcal{F}}$) applies PixelShuffle \cite{shi2016real} upsampling to part \emph{a}, and downsampling to part \emph{b}. Subsequently, parts \emph{a} and \emph{b} merge and undergo continuous convolution operations, eventually leading to the generation of bidirectional optical flow $\bm{\phi}_t$ and mask $\bm{\mu}_t$ corresponding to the specific $t$ through upsampling, as:

\begin{align}
\bm{\phi}_t, \bm{\mu}_t=\bm{\mathcal{F}} \left(\bm{\mathcal{C}}\left(\bm{M}_{0 \rightarrow t}, \bm{M}_{1 \rightarrow t}, \bm{S}_0, \bm{S}_1\right), \mathcal{\bm{\mathcal{C}}}\left(\bm{I}_0, \bm{I}_1\right)\right).
\end{align}

Next, we initially employ $\bm{\phi}_t$ to warp $\bm{I}_0$, $\bm{I}_1$, as well as the low-level features $\bm{L}_j^i$ from different layers extracted by FE, and the general structural features $\bm{S}_0$ and $\bm{S}_1$ extracted by MFSE. For instance, for $\bm{X}_z^y$, the result after warping is denoted as \scalebox{0.75}{$\widetilde{\bm{X}_z^y}$}. Subsequently, we concatenate $\bm{I}_0$, $\bm{I}_1$, \scalebox{0.75}{$\widetilde{\bm{I}_0}$}, \scalebox{0.75}{$\widetilde{\bm{I}_1}$}, $\bm{\phi}_t$, and $\bm{\mu}_t$ together, referred to as $\mathcal{O}_t$. 

\textbf{Refiner.} Finally, through the Refiner (a simplified UNet \cite{ronneberger2015u}), by integrating and refining features of different scales into $\mathcal{O}_t$ layer by layer, and then utilizing skip connection, we obtain the intermediate frame \scalebox{0.75}{$\widehat{\bm{I}_t}$} corresponding to $t$, as depicted in Fig. \ref{fig:arch}. Mathematically,

\begin{align}
\widehat{\bm{I}_t}=\widetilde{\bm{I}_t}+\bm{R}\left(\bm{\mathcal{O}}_t, \bm{L}, \bm{S}\right),
\end{align}
where $\bm{R}$ signifies the Refiner, $\bm{L}$ denotes the collection of $\bm{L}_j^i$, and $\bm{S}$ refers to the collection of $\bm{S}_0$ and $\bm{S}_1$. The symbol $\odot$ represents the Hadamard product, and \scalebox{0.75}{$\widetilde{\bm{I}_t}$} is determined as follows:

\begin{equation}
\begin{aligned}
\widetilde{\bm{I}_t}= & \ \bm{\mu}_t \odot \operatorname{backwarp}\left(\bm{I}_0, \bm{\phi}_{t \rightarrow 0}\right) \\
+ & \ (1-\bm{\mu}_t) \odot \operatorname{backwarp}\left(\bm{I}_1, \bm{\phi}_{t \rightarrow 1}\right).
\end{aligned}
\end{equation}

\begin{figure*}[ht]
    \centering
    \includegraphics[width = 1.0\linewidth]{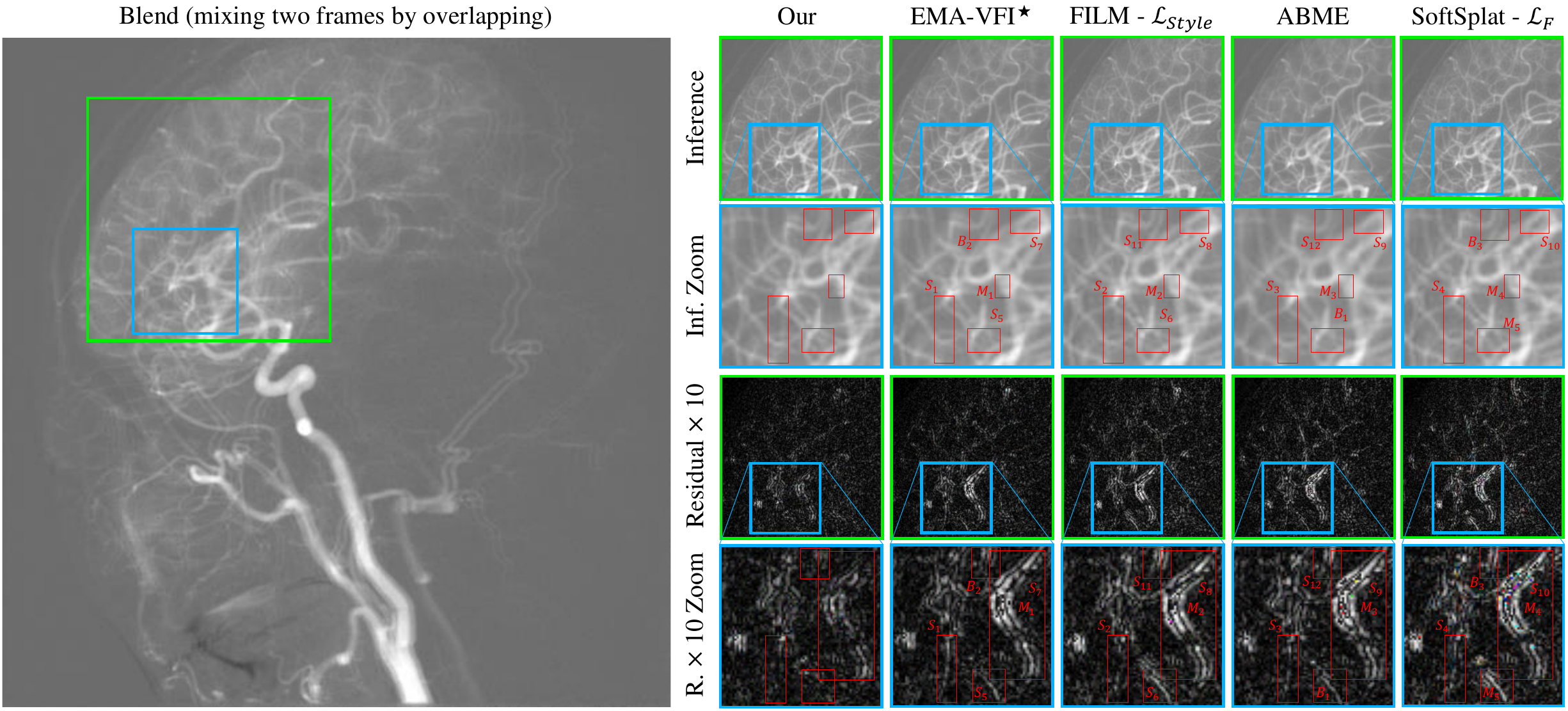}
    \vspace{-0.18in}
    \caption{\textbf{Visual comparison for interpolating one frame with methods in VFI.} "$\star$" indicates SOTA. On the right side, the first and third rows correspond to the \textbf{\textcolor{GrassGreen}{green}} box in blend, and the second and fourth rows correspond to the \textbf{\textcolor{SkyBlue}{blue}} box in blend. $M$ stands for motion artifact, $S$ for structural dissipation, and $B$ for blurring. Comparing the results of various methods, it can be proved that the natural scene VFI method has many problems of motion artifact, structural dissipation and blurring, while MoSt-DSA relatively obviously alleviates these problems.}
    \label{fig:com_sota_inf1}
    \vspace{-0.1in}
\end{figure*}

\subsection{Loss Functions}

To further enhance the inference quality, we employed a combination of three types of loss functions, as follows:

\begin{equation}
\mathcal{L}=\bm{w}_1 \mathcal{L}_1+\bm{w}_{\mathrm{VGG}} \mathcal{L}_{\mathrm{VGG}}+\bm{w}_{\mathrm{Style}} \mathcal{L}_{\mathrm{Style}},
\end{equation}
where $\mathcal{L}_1$ denotes the L1 reconstruction loss, which minimizes the pixel-wise RGB difference. Additionally, $\mathcal{L}_{\mathrm{VGG}}$ employs the L1 norm of the VGG-19 features to enhance finer image details and texture quality \cite{simonyan2014very}. The style loss $\mathcal{L}_{\mathrm{Style}}$ utilizes the L2 norm of the auto-correlation of the VGG-19 features \cite{gatys2016image,reda2018sdc,liu2018image}. This approach aims to further leverage the benefits of $\mathcal{L}_{\mathrm{VGG}}$ by capturing and replicating style patterns and textures more effectively. Regarding the selection of the weights ($\bm{w}_1$, $\bm{w}_{\mathrm{VGG}}$, $\bm{w}_{\mathrm{Style}}$), we referenced \cite{reda2022film}.

\section{Experiments}

\subsection{Datasets}

We collected 470 head DSA image sequences from 8 hospitals, each from a different patient, typically containing 152 images of 489x489 resolution. These were split into 329 for training and 141 for testing, maintaining a 7:3 ratio.
For each sequence targeting $n$-frame interpolation, we arrange it into several groups, each with consecutive $n+2$ frames. Adjacent groups start one frame apart. For details regarding data acquisition, we use NeuAngio33C, NeuAngio43C, and NeuAngio-CT equipment, following the SpinDSA protocol.

\subsection{Implementation Details}
\textbf{Model Configuration.} For interpolating 1 to 3 frames, time ($t$) sequences are set to [0.5], [0.33, 0.67], and [0.25, 0.50, 0.75], respectively. For simulating contextual interactions, context modeling scope ($r$) sizes are 29, 29, and 21. Effects of varying $r$ are compared in the ablation study.

\textbf{Training Details.} We trained on 4 A100 GPUs, and for tasks interpolating 1 to 3 frames, we set the batch sizes to 10, 10, and 6, with warm-up steps set to 9000, 12000, and 16000, respectively. We use the AdamW \cite{loshchilov2018fixing} optimizer with $\beta_1=0.9$, $\beta_2=0.999$, and a weight decay of $1e-4$. The learning rate is warmed up to $2e-4$ and then decays following a cosine schedule \cite{loshchilov2016sgdr}, decreasing to $2e-5$ over 300 epochs. We crop each frame to a resolution of 320 × 320 and apply random flip and rotation for augmentation. Regarding the selection of loss weights ($\bm{w}_1$, $\bm{w}_{\mathrm{VGG}}$, $\bm{w}_{\mathrm{Style}}$), we referenced \cite{reda2022film}, assigning weights of ($1.0, 1.0, 0.0$) for the first epoch and weights of ($1.0, 0.25, 40.0$) for the subsequent epochs.

\textbf{Testing Details.} To highlight our method's advantages, we compared MoSt-DSA with representative VFI methods. For ABME \cite{park2021abme} and SoftSplat \cite{niklaus2020softmax}, we tested on released pre-trained weights due to the absence of training codes. For EMA-VFI (state-of-the-art) \cite{zhang2023extracting} and FILM \cite{reda2022film}, we retrained them on our dataset following their original setups. All tests were performed on a single RTX 3090 GPU.

\textbf{Comparison Details.} We trained two versions: one (MoSt-DSA-$\mathcal{L}_1$) using only the $\mathcal{L}_1$ loss, which achieves higher test scores; the other (MoSt-DSA) using our proposed combined loss $\mathcal{L}$, which benefits image quality (see supplementary materials for proof). When comparing visual effects, we use the version of the model that yields high image quality\cite{reda2022film,gatys2016image}, i.e., FILM-$\mathcal{L}_{Style}$, and SoftSplat-$\mathcal{L}_F$.

\subsection{Single-Frame Interpolation}

\begin{table}[t]
  \centering
  \caption{\textbf{Quantitative comparison with VFI methods on \textbf{single-frame} interpolation.} Best scores for color losses in \textbf{\textcolor{blue}{blue}}, and for perceptually-sensitive losses in \textbf{\textcolor{red}{red}}. The second lowest memory usage in \textbf{\textcolor{GrassGreen}{green}}. "$\star$" indicates SOTA. EMA is short for EMA-VFI.}
  \scalebox{0.85}{
    \begin{tabular}{ccccccc}
    \toprule
    \multirow{2}[3]{*}{} & \multicolumn{2}{c}{\textbf{SSIM (\boldmath{$\%$})}} & \multicolumn{2}{c}{\textbf{PSNR}} & \textbf{Time (s)} & \textbf{Memory} \\
    \cmidrule{2-7}       & Mean$\uparrow$ & STD$\downarrow$  & Mean$\uparrow$ & STD$\downarrow$  & 1 frame$\downarrow$ & 1 frame$\downarrow$ \\
    \midrule
    ABME \cite{park2021abme} & 94.02  & 2.28 & 39.83 & 3.39  & 0.383  & 2.66 G  \\
    FILM-$\mathcal{L}_1$ \cite{reda2022film} & 94.11  & 2.37 & 39.86 & 3.43  & 0.201  & 3.51 G  \\
    EMA-small \cite{zhang2023extracting} & 94.19  & 2.21 & 40.07 & 3.47  & 0.027  & \textbf{\textcolor{blue}{2.10 G}}  \\
    EMA$^\star$ \cite{zhang2023extracting}  & 94.33  & 2.13 & 40.13 & 3.40  & 0.056  & 2.63 G  \\
    MoSt-DSA-$\mathcal{L}_1$ & \textbf{\textcolor{blue}{94.62}}  & \textbf{\textcolor{blue}{2.12}} & \textbf{\textcolor{blue}{40.32}} & \textbf{\textcolor{blue}{3.35}}  & \textbf{\textcolor{blue}{0.024}}  & \textbf{\textcolor{GrassGreen}{2.59 G}}  \\
    \midrule
    SoftSplat-$\mathcal{L}_F$ \cite{niklaus2020softmax} & 91.78  & 3.07 & 38.59 & 3.51  & 0.035  & \textbf{\textcolor{red}{2.20 G}}  \\
    FILM-$\mathcal{L}_{VGG}$ \cite{reda2022film} & 93.10  & 2.67 & 39.27 & 3.45  & 0.201  & 3.51 G  \\
    FILM-$\mathcal{L}_{Style}$ \cite{reda2022film} & 93.05  & 2.72 & 39.25 & 3.48  & 0.201  & 3.51 G  \\
    MoSt-DSA & \textbf{\textcolor{red}{93.65}}  & \textbf{\textcolor{red}{2.61}} & \textbf{\textcolor{red}{39.55}} & \textbf{\textcolor{red}{3.45}}  & \textbf{\textcolor{red}{0.024}}  & \textbf{\textcolor{GrassGreen}{2.59 G}}  \\
    \bottomrule
    \end{tabular}%
  }
  \label{tab:inf1_ave}%
\end{table}%

\begin{table}[t]
  \centering
  \caption{\textbf{Quantitative comparison with VFI methods on \textbf{two frames} interpolation.} The meanings of \textbf{\textcolor{blue}{blue}}, \textbf{\textcolor{red}{red}}, \textbf{\textcolor{GrassGreen}{green}}, and "$\star$" are the same as those in Table \ref{tab:inf1_ave}. EMA is short for EMA-VFI.}
  \scalebox{0.85}{
    \begin{tabular}{ccccccc}
    \toprule
    \multirow{2}[3]{*}{} & \multicolumn{2}{c}{\textbf{SSIM (\boldmath{$\%$})}} & \multicolumn{2}{c}{\textbf{PSNR}} & \textbf{Time (s)} & \textbf{Memory} \\
    \cmidrule{2-7}       & Mean$\uparrow$ & STD$\downarrow$  & Mean$\uparrow$ & STD$\downarrow$  & 2 frame$\downarrow$ & 2 frame$\downarrow$ \\
    \midrule
    FILM-$\mathcal{L}_1$ \cite{reda2022film} & 92.62 & 3.13 & 37.93 & 3.82  & 0.388  & 3.58 G \\
    EMA-small \cite{zhang2023extracting} & 91.82 & 3.68 & 37.37 & 4.07  & 0.074  & \textbf{\textcolor{blue}{2.10 G}} \\
    EMA$^\star$ \cite{zhang2023extracting}   & 91.90 & 3.63 & 37.41 & 4.08  & 0.112  & 2.64 G \\
    MoSt-DSA-$\mathcal{L}_1$ & \textbf{\textcolor{blue}{94.35}} & \textbf{\textcolor{blue}{2.29}} & \textbf{\textcolor{blue}{39.78}} & \textbf{\textcolor{blue}{3.44}} & \textbf{\textcolor{blue}{0.070}}  & \textbf{\textcolor{GrassGreen}{2.61 G}} \\
    \midrule
    SoftSplat-$\mathcal{L}_F$ \cite{niklaus2020softmax} & 90.86 & 3.49 & 37.84 & 3.43  & 0.084  & \textbf{\textcolor{red}{2.20 G}} \\
    FILM-$\mathcal{L}_{VGG}$ \cite{reda2022film} & 91.42 & 3.42 & 37.47 & 3.67  & 0.388  & 3.58 G \\
    FILM-$\mathcal{L}_{Style}$ \cite{reda2022film} & 91.31 & 3.50 & 37.39 & 3.74  & 0.388  & 3.58 G \\
    MoSt-DSA & \textbf{\textcolor{red}{93.14}} & \textbf{\textcolor{red}{2.84}} & \textbf{\textcolor{red}{38.94}} & \textbf{\textcolor{red}{3.38}} & \textbf{\textcolor{red}{0.070}}  & \textbf{\textcolor{GrassGreen}{2.61 G}} \\
    \bottomrule
    \end{tabular}%
  }
  \label{tab:inf2_ave}%
\end{table}%

\begin{table}[t]
  \centering
  \caption{\textbf{Quantitative comparison with VFI methods on \textbf{three frames} interpolation.} The meanings of \textbf{\textcolor{blue}{blue}}, \textbf{\textcolor{red}{red}}, \textbf{\textcolor{GrassGreen}{green}}, and "$\star$" are the same as those in Table \ref{tab:inf1_ave}. EMA is short for EMA-VFI.}
  \scalebox{0.85}{
    \begin{tabular}{ccccccc}
    \toprule
    \multirow{2}[3]{*}{} & \multicolumn{2}{c}{\textbf{SSIM (\boldmath{$\%$})}} & \multicolumn{2}{c}{\textbf{PSNR}} & \textbf{Time (s)} & \textbf{Memory} \\
    \cmidrule{2-7}       & Mean$\uparrow$ & STD$\downarrow$  & Mean$\uparrow$ & STD$\downarrow$  & 3 frame$\downarrow$ & 3 frame$\downarrow$ \\
    \midrule
    FILM-$\mathcal{L}_1$ \cite{reda2022film} & 91.94 & 3.52 & 37.21 & 3.91  & 0.548  & 3.58 G \\
    EMA-small \cite{zhang2023extracting} & 90.48 & 4.52 & 36.31 & 4.24  & 0.122  & \textbf{\textcolor{blue}{2.10 G}} \\
    EMA$^\star$ \cite{zhang2023extracting}   & 90.57 & 4.50 & 36.35 & 4.24  & 0.165  & 2.64 G \\
    MoSt-DSA-$\mathcal{L}_1$ & \textbf{\textcolor{blue}{93.58}} & \textbf{\textcolor{blue}{2.69}} & \textbf{\textcolor{blue}{38.85}} & \textbf{\textcolor{blue}{3.56}} & \textbf{\textcolor{blue}{0.117}}  & \textbf{\textcolor{GrassGreen}{2.61 G}} \\
    \midrule
    SoftSplat-$\mathcal{L}_F$ \cite{niklaus2020softmax} & 90.07 & 3.87 & 37.24 & \textbf{\textcolor{red}{3.53}} & 0.137  & \textbf{\textcolor{red}{2.20 G}} \\
    FILM-$\mathcal{L}_{VGG}$ \cite{reda2022film} & 90.63 & 3.83 & 36.75 & 3.74  & 0.548  & 3.58 G \\
    FILM-$\mathcal{L}_{Style}$ \cite{reda2022film} & 90.54 & 3.90 & 36.66 & 3.82  & 0.548  & 3.58 G \\
    MoSt-DSA & \textbf{\textcolor{red}{93.03}} & \textbf{\textcolor{red}{2.94}} & \textbf{\textcolor{red}{38.66}} & \textbf{\textcolor{GrassGreen}{3.59}} & \textbf{\textcolor{red}{0.117}}  & \textbf{\textcolor{GrassGreen}{2.61 G}} \\
    \bottomrule
    \end{tabular}%
  }
  \label{tab:inf3_ave}%
\end{table}%

\begin{figure}[t]
    \centering
    \includegraphics[width = 0.96\linewidth]{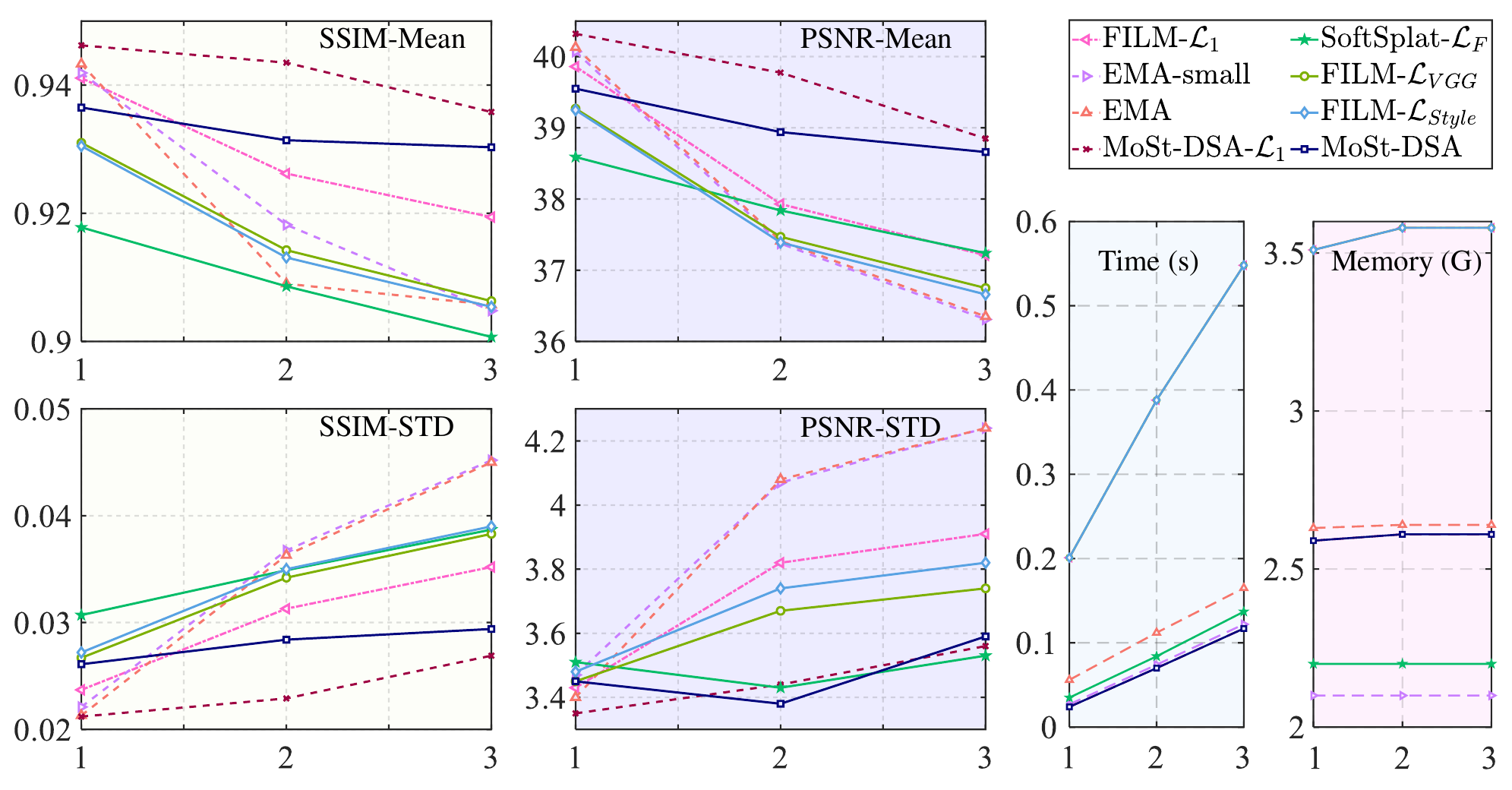}
    \vspace{-0.05in}
    \caption{\textbf{Intuitive comparison of metrics for each method interpolating 1 to 3 frames.} Our MoSt-DSA-$\mathcal{L}_1$ leads SOTA EMA-VFI in all respects, while showing superior robustness with a lower STD.}
    \label{fig:1to3_compa}
\end{figure}

\begin{figure*}[t]
    \centering
    \includegraphics[width = 1.0\linewidth]{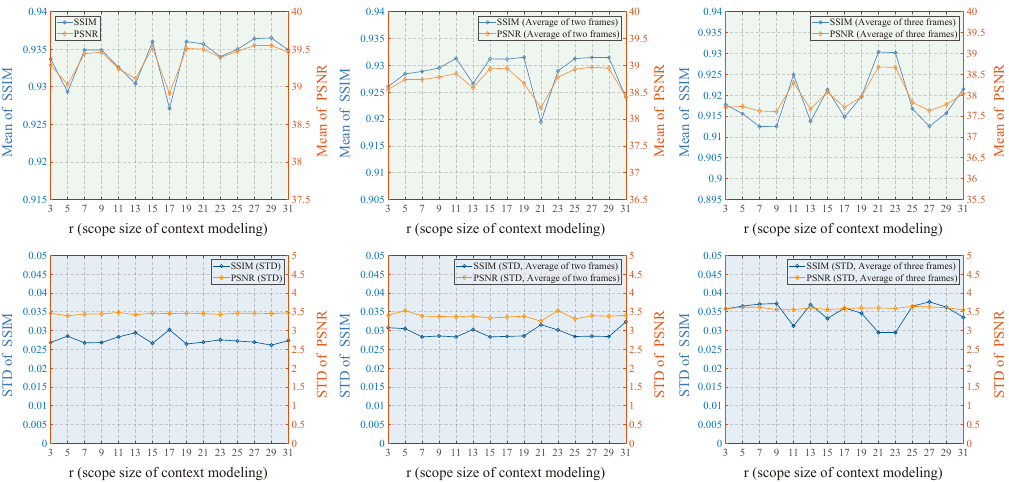}
    \vspace{-0.18in}
    \caption{\textbf{Impact of context modeling scope (\boldmath{$r$}) sizes for DSA frame interpolation tasks: from 1 to 3 frames.} The first to third columns correspond to interpolating 1 to 3 frames, and the first to second rows represent the mean and STD, respectively. The stable STD proves the robustness of MSFE, and the Mean indicates that the best $r$ for interpolating frames 1 to 3 is 29, 29, and 21, respectively.}
    \label{fig:ablation_r}
    \vspace{-0.05in}
\end{figure*}

We visualized the single-frame interpolation results of each model and compared them with the ground truth by calculating residuals.

As shown in Fig. \ref{fig:com_sota_inf1}, the first and third rows correspond to the green box in blend, and the second and fourth rows correspond to the blue box in blend. $M$ stands for motion artifact, $S$ for structural dissipation, and $B$ for blurring. By comparing the results of various methods, it can be proved that the natural scene VFI method has many problems of motion artifact, structural dissipation and blurring, while MoSt-DSA relatively obviously alleviates these problems.

The quantitative comparison for single-frame interpolation, as shown in Tab. \ref{tab:inf1_ave}, demonstrates our MoSt-DSA's superiority in SSIM, PSNR, and inference time over all competitors. Furthermore, our model also boasts more efficient memory usage than EMA-VFI during inference.
Notably, the score differences among SOTA methods in the VFI domain are minimal. For instance, on the UCF101 dataset \cite{soomro2012ucf101}, the top-performing EMA-VFI surpasses the second-best \cite{jin2023unified} by only 0.01\% in SSIM and 0.01 in PSNR, and the third-best \cite{zhou2023video} by 0.04\% in SSIM and 0.01 in PSNR. Thus, it is a significant margin that our MoSt-DSA-$\mathcal{L}_1$'s lead over the EMA-VFI, by \textbf{0.29\%} in SSIM and \textbf{0.19} in PSNR, as shown in Tab. \ref{tab:inf1_ave}.

\subsection{Direct Multi-Frame Interpolation}

We further compared our method with representative VFI methods in tasks of direct interpolating 2 and 3 frames.

For each method, we set t=[0.33, 0.67] and t=[0.25, 0.50, 0.75] for interpolating 2 to 3 frames, respectively. Considering that ABME \cite{park2021abme} couldn't interpolate at arbitrary time steps, we excluded it from the comparison. We give the average values across metrics for each frame count. For instance, if interpolating 2 frames results in SSIM values of [0.8, 0.9], then the average is 0.85.

The quantitative evaluation results for interpolating 2 and 3 frames are presented in Tab. \ref{tab:inf2_ave} and \ref{tab:inf3_ave}. Our MoSt-DSA continues to outperform other methods, in terms of SSIM, PSNR, and inference time, also exhibiting a lower standard deviation (STD). Memory usage during inference also remains more efficient than the EMA-VFI. This conclusively demonstrates the superior robustness of our method. 

We further intuitively compared the metrics for interpolating 1 to 3 frames, 
Fig. \ref{fig:1to3_compa} shows a clear pattern: MoSt-DSA's superiority in SSIM, PSNR, and stability grows with the increase in interpolated frames. Compared to EMA-VFI, MoSt-DSA-$\mathcal{L}_1$'s SSIM is higher by \textbf{0.29\%}, \textbf{2.45\%}, and \textbf{3.01\%} for interpolating 1, 2, and 3 frames. In terms of PSNR, the increase is \textbf{0.19}, \textbf{2.37}, and \textbf{2.50}. Furthermore, MoSt-DSA-$\mathcal{L}_1$'s STD for SSIM is lower by 0.4\%, 37\%, and 40\%, and for PSNR, it is 1.49\%, 16\%, and 16\% lower. We believe this significant lead reflects the advantages of MoSt-DSA trained with multi-frame supervision to model motion and structural interactions accurately, and highlights the importance of multi-frame supervision training for direct multi-frame interpolation tasks.

\subsection{3D Reconstruction Showcase from Single Frame Interpolation}

We conducted 3D reconstructions using both the single-frame interpolated sequences (interpolating every other image) and the original DSA sequences. Our results are virtually indistinguishable to the reconstruction from original data. Details in supplementary materials.

\subsection{Ablation Study}

\textbf{Impact of context modeling scope (\boldmath{$r$}).} Fig. \ref{fig:ablation_r} demonstrates that the impact of $r$ on STD is minimal, highlighting MSFE's robustness. $r$'s influence is slightly more pronounced on Mean-of-SSIM (no more than 1.8\%) than on Mean-of-PSNR (no more than 1.1). More detailed numerical results are available in the supplementary materials.                                                                

\textbf{Loss function comparison on our MoSt-DSA.} We prove that our proposed loss function significantly improves image quality, in the supplementary materials.

\section{Conclusion}

We have proposed MoSt-DSA, the first work that uses deep learning for DSA frame interpolation, to reduce radiation dose in DSA imaging significantly. In particular, we devised a general module that models motion and structural context interactions between frames in a fully convolutional manner, by adjusting the optimal context range and transforming available contexts into linear functions. Experiment results show that our MoSt-DSA outperforms the state-of-the-art Video Frame Interpolation methods in accuracy, speed, visual effect, and memory usage for interpolating 1 to 3 frames, and can also assist physicians in 3D diagnosis and treatment.


\newpage
\bibliography{paper}

\newpage
\section*{Supplementary Materials}
\subsection*{S. 1. Detailed Impact of Context Modeling Scope ($r$)}
Detailed numerical results are as Tab. \ref{tab:ablation-inf1-ave}, \ref{tab:ablation-inf2-ave}, and \ref{tab:ablation-inf3-ave}, consistent with the conclusions of \S 4.6 in the main document.

We would like to highlight the following points:

\begin{itemize}
\item Our MoSt-DSA does not significantly increase memory usage during multi-frame interpolation inference (consuming \textbf{2.59}, \textbf{2.61}, and \textbf{2.61G} for interpolating 1 to 3 frames, respectively), enabling convenient offline deployment on memory-constrained devices and assisting physicians in reducing DSA radiation exposure.
\item Furthermore, our MoSt-DSA maintains low computation times (\textbf{0.024}, \textbf{0.070}, and \textbf{0.117s} for interpolating 1 to 3 frames, respectively), saving valuable time for patient treatment. 
\end{itemize}

\begin{table}[htbp]
  \centering
  \caption{\textbf{Impact of context scope (\boldmath{$r$}) on \textbf{single frame} interpolation.} The best scores and the second best are in \textbf{\textcolor{red}{red}} and \textbf{\textcolor{blue}{blue}} respectively.}
  \scalebox{0.85}{
    \begin{tabular}{ccccccc}
    \toprule
    \multirow{2}[3]{*}{\boldmath{$r$}} & \multicolumn{2}{c}{\textbf{SSIM (\boldmath{$\%$)}}} & \multicolumn{2}{c}{\textbf{PSNR}} & \textbf{Time (s)} & \textbf{Memory} \\
    \cmidrule{2-7}        & Mean$\uparrow$ & STD$\downarrow$  & Mean$\uparrow$ & STD$\downarrow$ & 1 frame$\downarrow$ & 1 frame$\downarrow$ \\
    \midrule
    3     & 93.37  & 2.68  & 39.29  & 3.46  & 0.024  & 2.59 G \\
    5     & 92.93  & 2.85  & 39.03  & \textbf{\textcolor{red}{3.39}}  & 0.025  & 2.59 G \\
    7     & 93.49  & 2.67  & 39.44  & 3.44  & 0.025  & 2.59 G \\
    9     & 93.49  & 2.68  & 39.46  & 3.44  & 0.024  & 2.59 G \\
    11    & 93.26  & 2.83  & 39.24  & 3.48  & 0.025  & 2.59 G \\
    13    & 93.04  & 2.94  & 39.11  & \textbf{\textcolor{blue}{3.42}}  & 0.025  & 2.59 G \\
    15    & 93.60  & 2.66  & \textbf{\textcolor{blue}{39.52}}  & 3.46  & 0.025  & 2.59 G \\
    17    & 92.71  & 3.02  & 38.91  & 3.45  & 0.025  & 2.59 G \\
    19    & 93.60  & \textbf{\textcolor{blue}{2.64}}  & 39.51  & 3.46  & 0.025  & 2.59 G \\
    21    & 93.57  & 2.69  & 39.50  & 3.45  & 0.025  & 2.59 G \\
    23    & 93.40  & 2.75  & 39.39  & 3.43  & 0.024  & 2.59 G \\
    25    & 93.50  & 2.72  & 39.47  & 3.46  & 0.025  & 2.59 G \\
    27    & \textbf{\textcolor{blue}{93.64}}  & 2.69  & \textbf{\textcolor{red}{39.55}}  & 3.46  & 0.025  & 2.59 G \\
    29    & \textbf{\textcolor{red}{93.65}}  & \textbf{\textcolor{red}{2.61}}  & \textbf{\textcolor{red}{39.55}}  & 3.45  & 0.024  & 2.59 G \\
    31    & 93.49  & 2.73  & 39.46  & 3.47  & 0.024  & 2.59 G \\
    \bottomrule
    \end{tabular}%
  }
  \label{tab:ablation-inf1-ave}%
\end{table}%

\begin{table}[htbp]
  \centering
  \caption{\textbf{Impact of context scope (\boldmath{$r$}) on \textbf{two frames} interpolation.} The best scores and the second best are in \textbf{\textcolor{red}{red}} and \textbf{\textcolor{blue}{blue}} respectively.}
  \scalebox{0.85}{
    \begin{tabular}{ccccccc}
    \toprule
    \multirow{2}[3]{*}{\boldmath{$r$}} & \multicolumn{2}{c}{\textbf{SSIM (\boldmath{$\%$})}} & \multicolumn{2}{c}{\textbf{PSNR}} & \textbf{Time (s)} & \textbf{Memory} \\
    \cmidrule{2-7}        & Mean$\uparrow$ & STD$\downarrow$  & Mean$\uparrow$ & STD$\downarrow$ & 2 frames$\downarrow$ & 2 frames$\downarrow$ \\
    \midrule
    3     & 92.60 & 3.07 & 38.55 & 3.40 & 0.071  & 2.61 G \\
    5     & 92.84 & 3.05 & 38.73 & 3.53 & 0.071  & 2.61 G \\
    7     & 92.88 & \textbf{\textcolor{red}{2.83}} & 38.73 & 3.38 & 0.071  & 2.61 G \\
    9     & 92.95 & 2.85 & 38.78 & 3.37 & 0.071  & 2.61 G \\
    11    & 93.13 & \textbf{\textcolor{red}{2.83}} & 38.84 & 3.36 & 0.071  & 2.61 G \\
    13    & 92.66 & 3.02 & 38.58 & 3.38 & 0.071  & 2.61 G \\
    15    & 93.12 & \textbf{\textcolor{red}{2.83}} & 38.93 & 3.33 & 0.071  & 2.61 G \\
    17    & 93.11 & \textbf{\textcolor{blue}{2.84}} & \textbf{\textcolor{blue}{38.94}} & 3.36 & 0.071  & 2.61 G \\
    19    & \textbf{\textcolor{red}{93.15}} & 2.85 & 38.66 & 3.38 & 0.071  & 2.61 G \\
    21    & 91.94 & 3.15 & 38.20 & \textbf{\textcolor{red}{3.25}} & 0.071  & 2.61 G \\
    23    & 92.89 & 3.01 & 38.78 & 3.53 & 0.071  & 2.61 G \\
    25    & 93.12 & \textbf{\textcolor{blue}{2.84}} & 38.92 & \textbf{\textcolor{blue}{3.31}} & 0.070  & 2.61 G \\
    27    & \textbf{\textcolor{blue}{93.14}} & 2.85 & \textbf{\textcolor{red}{38.96}} & 3.39 & 0.070  & 2.61 G \\
    29    & \textbf{\textcolor{blue}{93.14}} & \textbf{\textcolor{blue}{2.84}} & \textbf{\textcolor{blue}{38.94}} & 3.38 & 0.070  & 2.61 G \\
    31    & 92.40 & 3.23 & 38.41 & 3.40 & 0.070  & 2.61 G \\
    \bottomrule
    \end{tabular}%
  }

  \label{tab:ablation-inf2-ave}%
\end{table}%

\begin{table}[t]
  \centering
  \caption{\textbf{Impact of context scope (\boldmath{$r$}) on \textbf{three frames} interpolation.} The best scores and the second best are in \textbf{\textcolor{red}{red}} and \textbf{\textcolor{blue}{blue}} respectively.}
  \scalebox{0.85}{
    \begin{tabular}{ccccccc}
    \toprule
    \multirow{2}[3]{*}{\boldmath{$r$}} & \multicolumn{2}{c}{\textbf{SSIM (\boldmath{$\%$})}} & \multicolumn{2}{c}{\textbf{PSNR}} & \textbf{Time (s)} & \textbf{Memory} \\
    \cmidrule{2-7}        & Mean$\uparrow$ & STD$\downarrow$  & Mean$\uparrow$ & STD$\downarrow$ & 3 frames$\downarrow$ & 3 frames$\downarrow$ \\
    \midrule
    3     & 91.77 & 3.57 & 37.70 & 3.58 & 0.118  & 2.61 G \\
    5     & 91.55 & 3.65 & 37.72 & 3.61 & 0.120  & 2.61 G \\
    7     & 91.24 & 3.70 & 37.62 & 3.62 & 0.118  & 2.61 G \\
    9     & 91.25 & 3.71 & 37.60 & 3.56 & 0.118  & 2.61 G \\
    11    & 92.49 & \textbf{\textcolor{blue}{3.12}} & 38.30 & \textbf{\textcolor{blue}{3.55}} & 0.117  & 2.61 G \\
    13    & 91.37 & 3.68 & 37.66 & 3.60 & 0.117  & 2.61 G \\
    15    & 92.13 & 3.32 & 38.07 & 3.56 & 0.117  & 2.61 G \\
    17    & 91.47 & 3.60 & 37.70 & 3.58 & 0.118  & 2.61 G \\
    19    & 91.96 & 3.46 & 37.95 & 3.60 & 0.117  & 2.61 G \\
    21    & \textbf{\textcolor{red}{93.03}} & \textbf{\textcolor{red}{2.94}} & \textbf{\textcolor{red}{38.67}} & 3.60 & 0.117  & 2.61 G \\
    23    & \textbf{\textcolor{blue}{93.01}} & \textbf{\textcolor{red}{2.94}} & \textbf{\textcolor{blue}{38.66}} & 3.58 & 0.118  & 2.61 G \\
    25    & 91.67 & 3.64 & 37.82 & 3.64 & 0.118  & 2.61 G \\
    27    & 91.25 & 3.76 & 37.62 & 3.63 & 0.118  & 2.61 G \\
    29    & 91.57 & 3.62 & 37.77 & 3.60 & 0.118  & 2.61 G \\
    31    & 92.14 & 3.35 & 38.03 & \textbf{\textcolor{red}{3.54}} & 0.117  & 2.61 G \\
    \bottomrule
    \end{tabular}%
  }
  \label{tab:ablation-inf3-ave}%
\end{table}%

\begin{figure}[t]
    \centering
    \includegraphics[width = 1.0\linewidth]{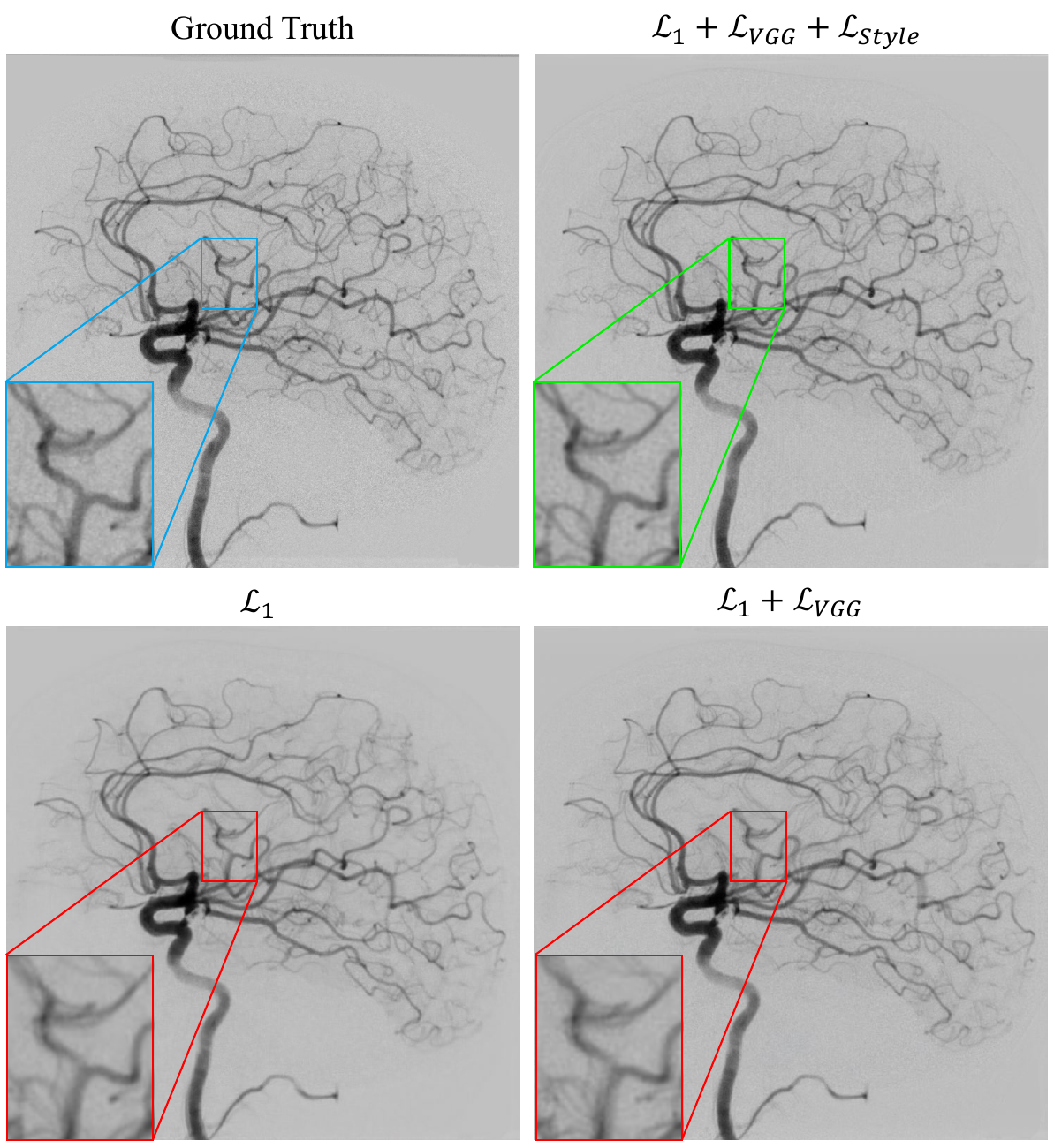}
    \caption{\textbf{Loss function comparison on our MoSt-DSA.} Our loss function strategy shows a significant improvement (\textbf{\textcolor{FluoGreen}{green}} boxes).}
    \label{fig:ablation_losses}
\end{figure}

\subsection*{S. 2. Loss Function Comparison on Our MoSt-DSA}
As shown in Fig. \ref{fig:ablation_losses}, solely using $\mathcal{L}_1$, or combining $\mathcal{L}_1$ with $\mathcal{L}_{\mathrm{VGG}}$, results in blurred image details (\textbf{\textcolor{red}{red}} boxes), whereas our combined loss function not only eliminates the blurriness but also closely approximates the ground truth (\textbf{\textcolor{FluoGreen}{green}} boxes).

\subsection*{S. 3. 3D Reconstruction Showcase from Single Frame Interpolation}

\begin{figure*}[htbp]
    \centering
    \includegraphics[width = 1.0\linewidth]{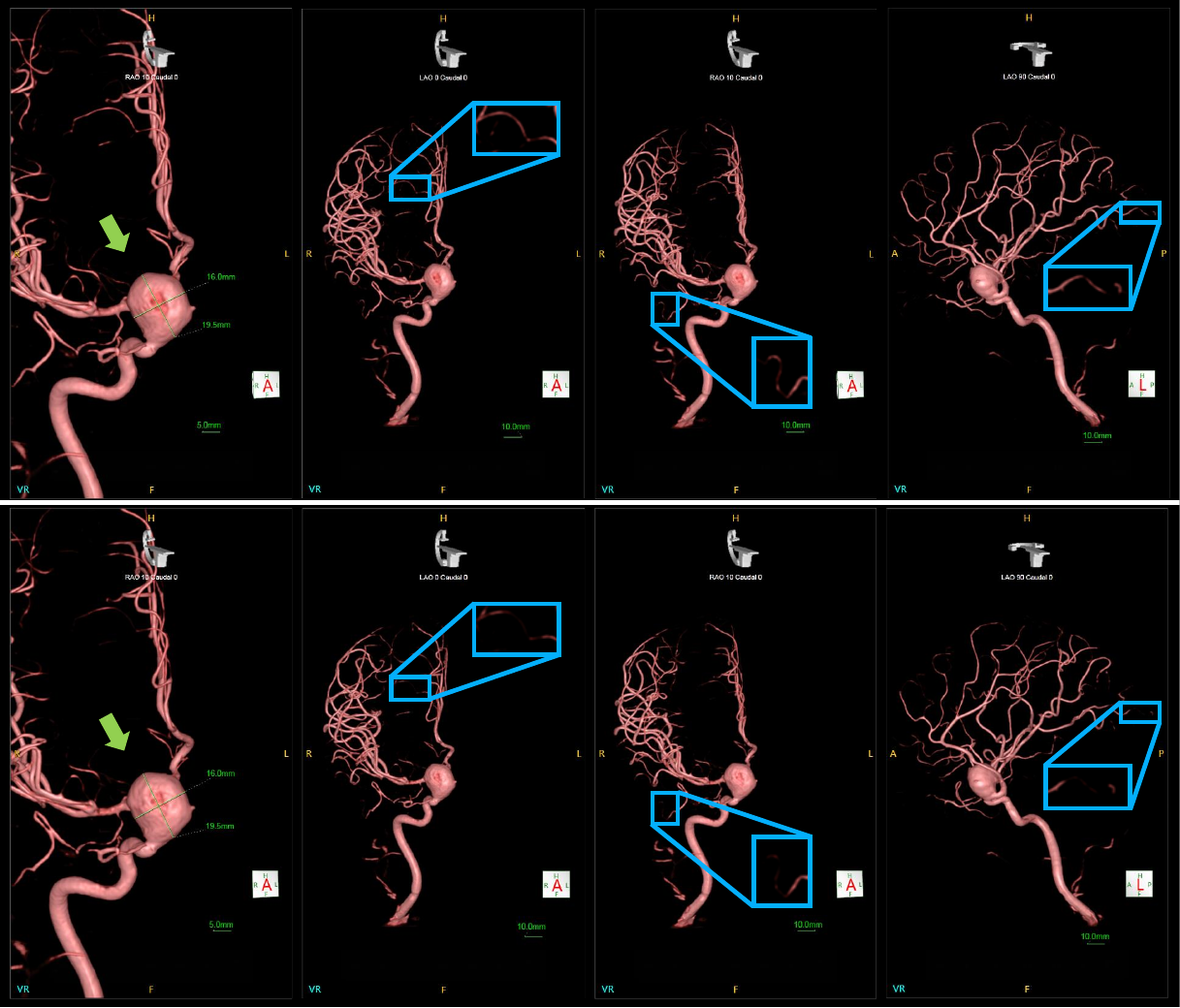}
    \caption{\textbf{3D reconstruction showcase: based on our single-frame interpolation result (below) vs. original data (above).} Our results are virtually indistinguishable to the reconstruction from original data, with differences only at the most delicate parts of the blood vessels (\textbf{\textcolor{SkyBlue}{blue}} boxes). Also, our results accurately maintain lesion sizes and details, aligning with the original data (\textbf{\textcolor{GrassGreen}{green}} arrows).}
    \label{fig:3d_reconstruction_larger}
\end{figure*}

We provide a showcase as Fig. \ref{fig:3d_reconstruction_larger}, consistent with the conclusions of \S 4.5 in the main document. As shown in Fig. \ref{fig:3d_reconstruction_larger}, our results are virtually indistinguishable to the reconstruction from original data, with differences only at the most delicate parts of the blood vessels (\textbf{\textcolor{SkyBlue}{blue}} box). Also, our results accurately maintain lesion sizes and details, aligning with the original data (\textbf{\textcolor{GrassGreen}{green}} arrows). The above conclusion indicates that our MoSt-DSA can also assist physicians in 3D diagnosis and treatment.

\end{document}